\documentclass[runningheads]{llncs}
\usepackage{graphicx}

\usepackage{amssymb}
\usepackage{amsmath}
\usepackage{mathabx}
\usepackage{mathabx}
\usepackage{mdwmath}
\usepackage{mdwtab}
\usepackage{eqparbox}
\usepackage{siunitx}
\usepackage{booktabs}
\usepackage{etoolbox}
\usepackage{cite}

\usepackage{xcolor}
\usepackage{hyperref}
\hypersetup{
    colorlinks=true,
    linkcolor=blue,
    filecolor=magenta,      
    urlcolor=cyan,
    citecolor=darkgray,
    breaklinks=true,
    pdftitle={The Enigma of Complexity},
    pdfpagemode=FullScreen,
}

\graphicspath{{Figures/}}

\begin{document}
\title{Quality-diversity for aesthetic evolution} 
%
%
\author{Jon McCormack\inst{1}\orcidID{0000-0001-6328-5064} \and
Camilo Cruz Gambardella \inst{1}\orcidID{0000-0002-8245-6778}}
\authorrunning{J. McCormack and C. Cruz Gambardella}
%
\institute{SensiLab, Monash University, Melbourne, Australia\\
\email{Jon.McCormack@monash.edu} \email{Camilo.CruzGambardella@monash.edu}\\
\url{https://sensilab.monash.edu}}

\maketitle              
\begin{abstract}
Many creative generative design spaces contain multiple regions with individuals of high aesthetic value. Yet traditional evolutionary computing methods typically focus on optimisation, searching for the fittest individual in a population. In this paper we apply quality-diversity search methods to explore a creative generative system (an agent-based line drawing model). We perform a random sampling of genotype space and use individual artist-assigned evaluations of aesthetic quality to formulate a computable fitness measure specific to the artist and this system. To compute diversity we use a convolutional neural network to discriminate features that are dimensionally reduced into two dimensions. We show that the quality-diversity search is able to find multiple phenotypes of high aesthetic value. These phenotypes show greater diversity and quality than those the artist was able to find using manual search methods.


\end{abstract}
%
%
%
\section{Introduction}
\label{s:introduction}

A long standing challenge in creative evolutionary applications has been to find suitable fitness measures \cite{McCormack2005a}, particularly when such measures need to consider `subjective' elements, such as visual aesthetics and personal taste. While significant progress has been made in quantifying aesthetics and understanding human aesthetic judgement \cite{Leder2014,Johnson2019}, evolutionary computing methods typically focus on optimisation: finding the fittest individual in a population. However, in many art and design applications there is no single, best design, rather a variety of possibilities that can be considered of interest. Moreover, complex generative systems can present a design space that is broad and unexplored, with many so-called `Klondike spaces' \cite{Perkins96,Perkins2001} of creative gold that are hidden amongst the vast regions of the ordinary or uninteresting. Rather than a single optimum or best design, such systems may have \emph{many} designs that the designer would consider worthwhile.

Typically the way around this problem would be to use human-in-the-loop methods, such as the Interactive Genetic Algorithm (IGA), which substitutes formalised fitness measures for human aesthetic judgement \cite{Bentley1999}. However the limitations of this approach are well known \cite{Takagi2001}: human evaluation creates a bottleneck; subjective comparison is only possible for a small number of individuals (i.e.\ $< 20$); human users become fatigued after only a few generations; evolving to specific targets is often difficult or impossible; design space exploration without a goal is largely equivalent to a random walk. There have been many attempts to overcome these problems, e.g.~distributed evolution with multiple users \cite{Secretan2011}, but distributed techniques are obviously incompatible with individual designers or personal aesthetic preferences. 

In this paper we present a method suited to the automated generation of diverse landscapes of creative alternatives, using the principles of \emph{quality-diversity search} (QD-search, \cite{Pugh2016}). QD-search methods attempt to find a diverse range of high fitness individuals and have shown good success in a variety of domains \cite{gomes2015devising}, such as robotics \cite{tarapore2016different, pugh2015confronting, Pugh2016} and the generation of content for video games \cite{khalifa2018talakat}. For creative applications, the key challenge is in finding good, suitable, and independent measures of both quality and diversity. While our application focuses on a generative art system that produces line drawings suitable for physical plotting, the methodology presented here can apply to any visual art or design system. Our approach is based on the following assumptions:

\begin{itemize}
    \item There is a creative system that can generate two-dimensional (2D) visual images (phenotypes) from some supplied parameters (genotypes). 
    \item The format, quantity and order of the parameters is arbitrary.
    \item There is no restriction on the type or style of images produced, except that they can be represented as pixel-based 2D images.
    \item The system designer or artist is able to evaluate the aesthetic value or quality of the produced images (phenotypes).
    \item The creative aspects of the system are expressed by the images produced, i.e.~there are no external factors or conditions that determine quality or difference in the produced phenotypes.
\end{itemize}

These are standard ways that an artist or designer would work with a generative system. While we restrict our study to 2D images, other formats, such as 3D forms could be easily accommodated by rendering the 3D form as a 2D image for evaluation (e.g. as in ~\cite{McCormackLomas2020b}).

To determine quality we first generate a random sample of phenotypes and ask the artist to evaluate their aesthetic quality manually, using this evaluation to derive a measure of quality specific to the artist and the system. Based on previous studies, which have suggested image metrics can serve as a good proxy for personal aesthetics \cite{McCormackEnigma2021}, we analyse the aesthetically ranked phenotypes, computing various measures of complexity and image morphology, looking for high correlations between these measures and the artist-assigned measures of quality. This analysis allows us to formulate a computable measure of quality specific to the artist and the generative system.

Diversity can take a number of forms and hence, measures. As this study is focused on visual images, we consider diversity exclusively in the visual sense, i.e.~ identifying the range of visually distinct features that the system can generate. To measure this diversity computationally, we turn to neural networks trained to visually discriminate image features. Over the last few years Convolutional Neural Networks (CNNs) trained on very large image corpora have become highly successful at feature classification and object recognition -- on par with human evaluation. Performing classification requires identifying the salient features of an image. For our system we used the ResNet-152 classifier, removing the last four layers of the network to expose a feature vector as the network's output. As this vector is large (2048 elements), we then employ dimension reduction methods to compute a final, 2-dimensional diversity measure.

Together, these computable measures of quality (aesthetic fitness) and diversity (visual features) allow us to generate a variety of forms that the artist should find interesting using QD search. Once found, these individuals can be further fine-tuned manually or accepted as successful products of the evolutionary generative system. Our results show that we were able to find a number of highly fit individuals that the artist had not been able to locate using other methods.

Before describing the system, experiments and method in detail, we provide a brief review and explanation of QD-Search and its application to visual creation.

\section{Quality-diversity search}

The use of evolutionary computing methods for design exploration has been gaining traction for the past 10 years. Lehman and Stanley \cite{lehman2008exploiting} pioneered an approach that proposes a departure from fitness-driven optimisation, instead looking to find the `best' pathway through the search space. Under this premise, they developed a series of experiments focusing on novelty and diversity \cite{lehman2011abandoning}, rather than on a fitness function that describes the expected performance of the populations being evolved, which enabled them to `illuminate' areas of the search space that hadn't been revealed through optimisation-based approaches. Moreover, some of the solutions found in these previously unexplored areas of the search space proved to be as fit -- if not fitter -- than those found using traditional methods. This approach, they argue, is well suited for use in contexts in which either there is more than one optimal solution, or where objectives are not explicit (e.g. design and art).

Since the introduction of the novelty search principles in 2008 (see \cite{lehman2008exploiting}), a wide range of methods that use them has been developed, incorporating different ways of measuring novelty, various combinations of diversity and fitness, and multiple underlying evolutionary algorithms to drive the search process \cite{gomes2015devising}. For the work presented in this paper, we adopted an approach that combines optimisation and diversity (QD-search), as it bears resemblance to the way that human creative processes unfolds: generate a series of candidate solutions/objects/things, assess them under the light of what they are expected to be, but also in search of surprising elements, or as Alexander puts it, things that ``display new physical order'' \cite[p.1]{Alexander1964}. Select the `best' ones -- if there are any -- and use them as starting point to repeat the process.

The MAP-Elites algorithm developed by \cite{mouret2015illuminating}  can be understood as a similar process. The goal is to evolve a landscape of diverse, highly fit individuals. To achieve this the algorithm requires a measure of fitness, as well as a way of classifying individuals into categories that account for their diversity. Interesting creative applications of this particular algorithm can be found in \cite{Colton2021}, where it is used in conjunction with generative adversarial networks (GANs) to evolve image style transfer blends, while trying to maintain the resemblance between the original image and the transformed one, as well as in the work developed by \cite{khalifa2018talakat}, where a constrained version of MAP-Elites is used to evolve bullet patterns of different levels of difficulty for video games.

\section{Generative System}
\label{s:generativeSystem}
\begin{figure}
    \centering
\includegraphics[width=0.7\textwidth]{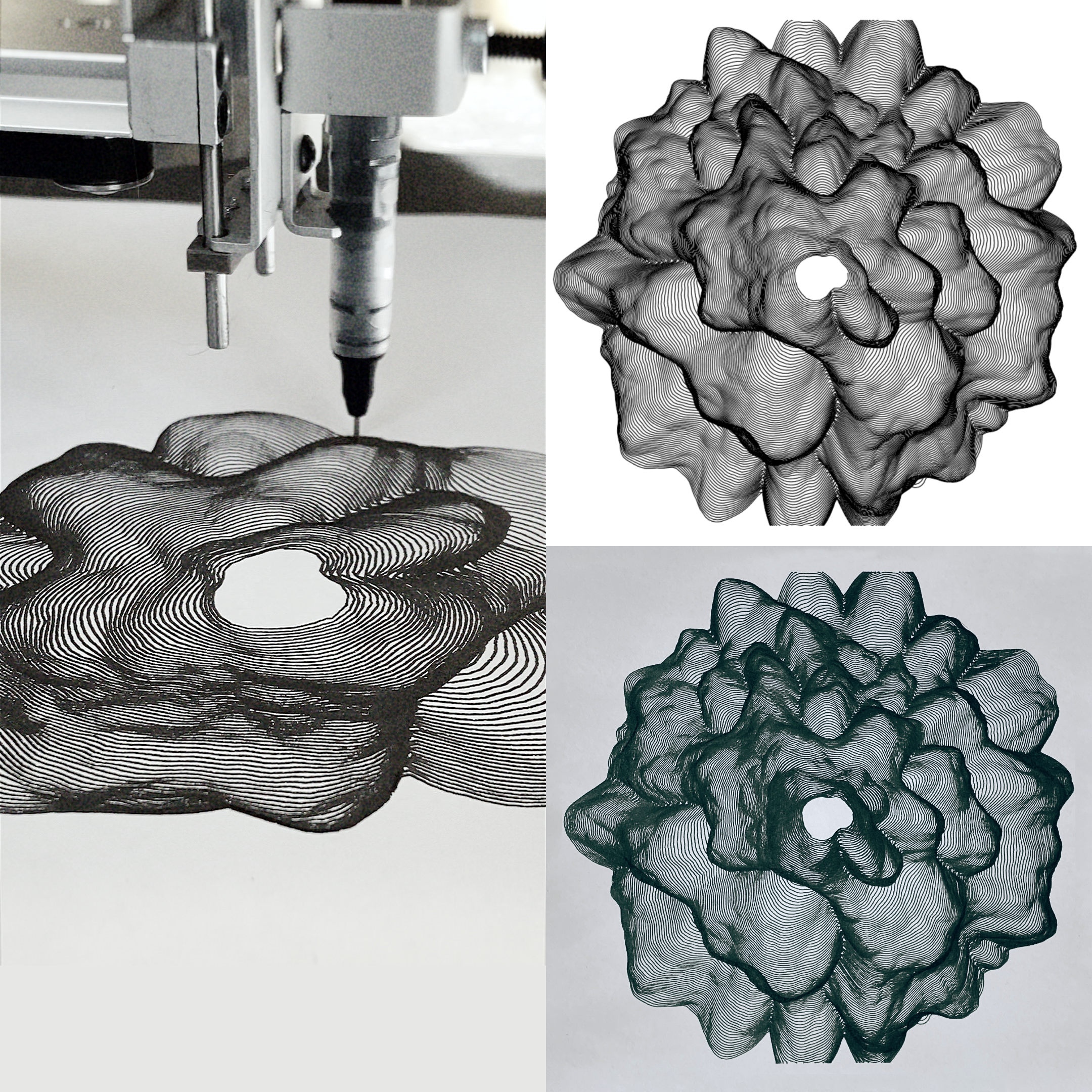}
    \caption{A sample drawing. The software outputs an svg vector image (top right), which is then fed to a pen plotter that traces out the image onto paper (left). The final resultant drawing (bottom right) takes about 30 mins to plot. }
    \label{fig:sample}
\end{figure}

To test the suitability of QD-search in creative applications, we developed a generative art system
using an agent-based line drawing model. 
Similar models have been effectively used in previous research, in addition to being recognised as artistically successful \cite{Annunziato1998,Baker1994,McCormack2010}.
A series of mobile agents are released onto a virtual ``canvas'' and proceed to draw a trail over their fixed lifetimes. The drawing is complete when all the agents are exhausted. The cumulative paths are output as an svg file, to facilitate high quality plotting. We use a line drawing system as the intended output is physical drawings, plotted with various permanent ink markers on paper. An example svg image and plotted drawing is shown in Fig. \ref{fig:sample}.

The movement of each agent is determined by a series of summed noise functions, based on Perlin noise \cite{Perlin2002} and procedural fluid flow \cite{Bridson2007}.
The system uses 17 different parameters (Table \ref{tab:genes}) to control the drawing, represented as normalised continuous real values in the range $[0-1]$. Parameters affect properties such as noise type ($g_{17}$) and frequency ($g_6$ \& $g_7$), number of drawing agents ($g_3$), the pen type ($g_{12}$ \& $g_{13}$), horizontal, vertical or circular pathways ($g_{14},g_{15},g_{16}$), agent speed ($g_2$) and lifetime ($g_3$). Each normalised parameter, $g_i$ is mapped to a parameter specific range, for example $g_1$ represents the width in pixels of the border from the edge of the canvas for agent placement and is mapped to the range $[0-400]$. The noise algorithm gene ($g_{17}$) maps to a set of five different discrete possibilities.

\begin{table}
    \centering
    \begin{tabular}{ll|ll}
      \textit{Gene}   &  \textit{Description} & \textit{Gene} & \textit{Description}\\
      \hline
      $g_1$ & border width & $g_{10}$ & noise octaves \\
      $g_2$ & agent speed & $g_{11}$ & noise falloff \\
      $g_3$ & agent density  & $g_{12}$ & agent/pen count \\
      $g_4$ & noise strength & $g_{13}$ & agent/pen ratio \\
      $g_5$ & noise displacement & $g_{14}$ & linear drawing style \\
      $g_6$ & noise x-frequency & $g_{15}$ & circular drawing style \\
      $g_7$ & noise y-frequency & $g_{16}$ & spiral drawing style \\
      $g_8$ & noise z-scale & $g_{17}$ & noise algorithm \\
      $g_9$ & z-position & & \\
      \hline \\
    
    \end{tabular}
    \caption{Genes used in the generative drawing system}
    \label{tab:genes}
\end{table}

\begin{figure}[!t]
    \centering
\includegraphics[width=\textwidth]{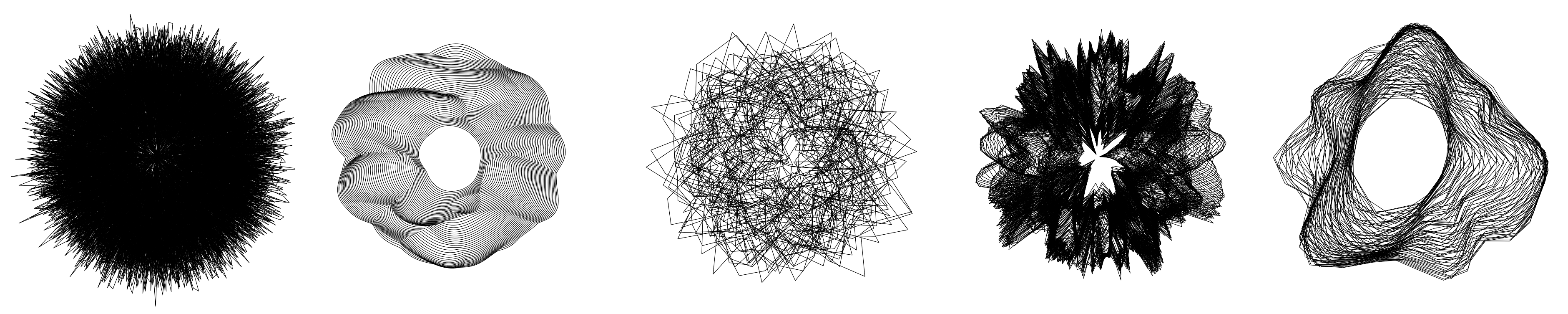}
    \caption{Five sample phenotypes selected from a pool of random genotypes. }
    \label{fig:five}
\end{figure}

Together these gene parameters ($g_i, i = 1\ldotp \ldotp17$) form a complete genotype ($G$) which deterministically maps to a phenotype ($G \rightarrow P$) by simulating the agents moving over the canvas using the parameters specified in $G$. There is no randomness in the generation, so any individual geneotype, $G_k$ will always produce the same equivalent phenotype, $P_k$. Fig. \ref{fig:five} shows some example phenotypes selected from a test set of randomly initialised genotypes. This test set was used to determine a computable fitness measure for the system, detailed next.

\section{Evolutionary method}
\label{s:evomethod}

\subsection{Fitness}
\label{ss:fitness}

Determining the aesthetic `quality' of an artwork is complex \cite{DBLP:conf/evoW/McCormack13}, as there are multiple factors that will influence how it is perceived, interpreted and ultimately appreciated \cite{Leder2014}. However, defining some measure of quality is crucial in the context of generative evolutionary art, as it is the `rein control'\cite{Clynes1969,Harvey2004} the artist needs to steer the generative process towards the results they are looking for. 

Previous work has shown that computable measures are far better at capturing an individual artist's perception of aesthetics in a specific system over collective general perceptions of quality or aesthetic value\cite{McCormackEnigma2021}. Relatively simple measures, such as image complexity \cite{Machado98, machado2015} or information content \cite{Moles1966} have been used as fitness measures, for example. More complex systems, such as deep learning classifiers, have also demonstrated good success in capturing an individual artist's concept of aesthetic quality \cite{McCormackLomas2020}.

\subsubsection{Human Fitness Evaluation}
Building on our previous results \cite{McCormackEnigma2021}, for this work we wanted to capture the artist's aesthetic preferences for the generative system in a formalised fitness measure. To do this we first generated a number of random phenotypes and asked the artist to evaluate their aesthetic quality using two different methods: direct numerical score and pairwise comparison ranking. The size of this dataset (257 images) was chosen as a compromise between getting a reasonable sampling of the design space and the overall fatigue and time required for the artist to perform both evaluation methods. The full dataset is available for download \cite{McCormack2022b}. Using two different evaluation methods allowed us to compare the pros and cons of each method, along with providing an understanding of the consistency of subjective evaluation. 

\begin{figure}
    \centering
    \begin{tabular}{c|c}
     \includegraphics[width=0.49\textwidth]{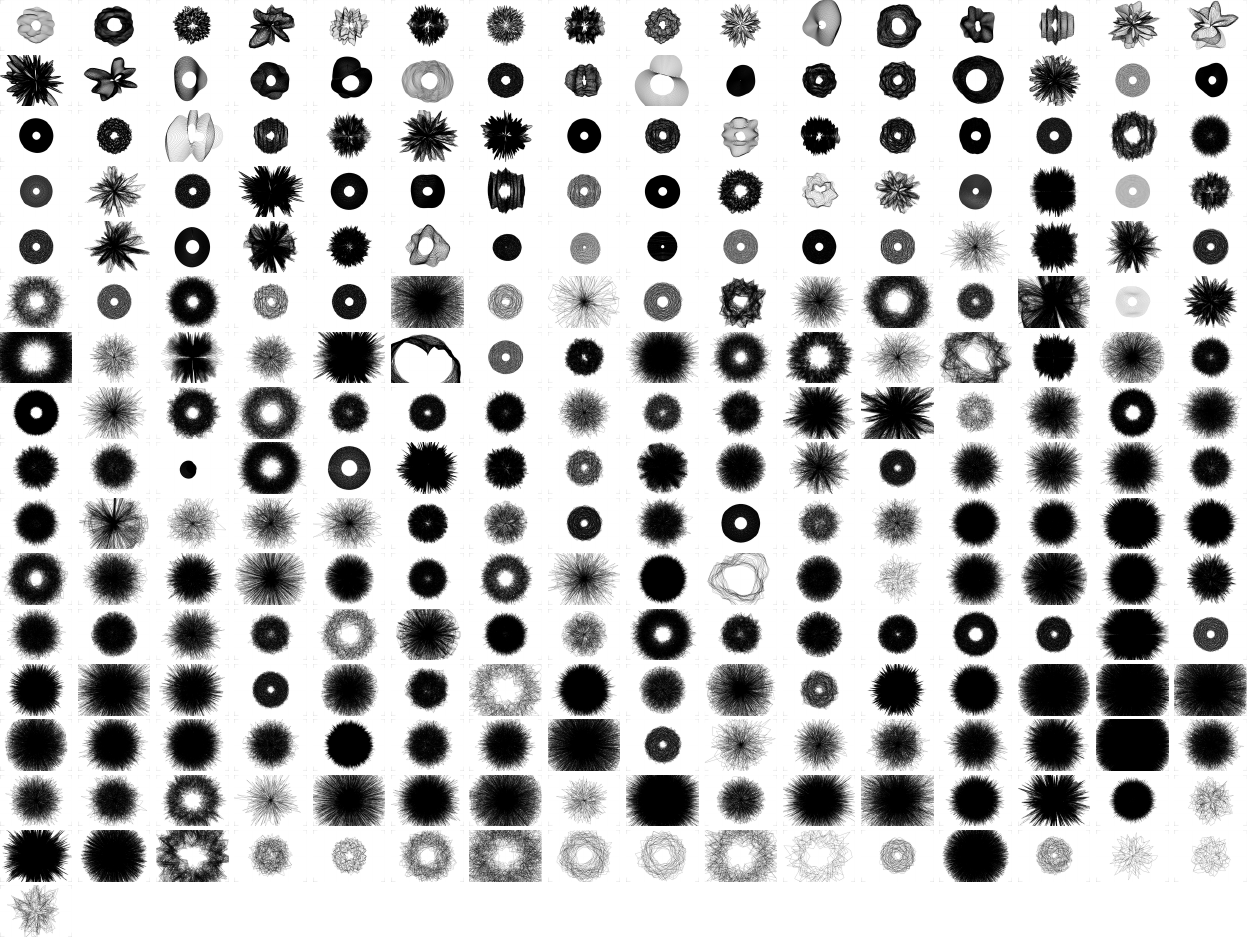} &
     \includegraphics[width=0.49\textwidth]{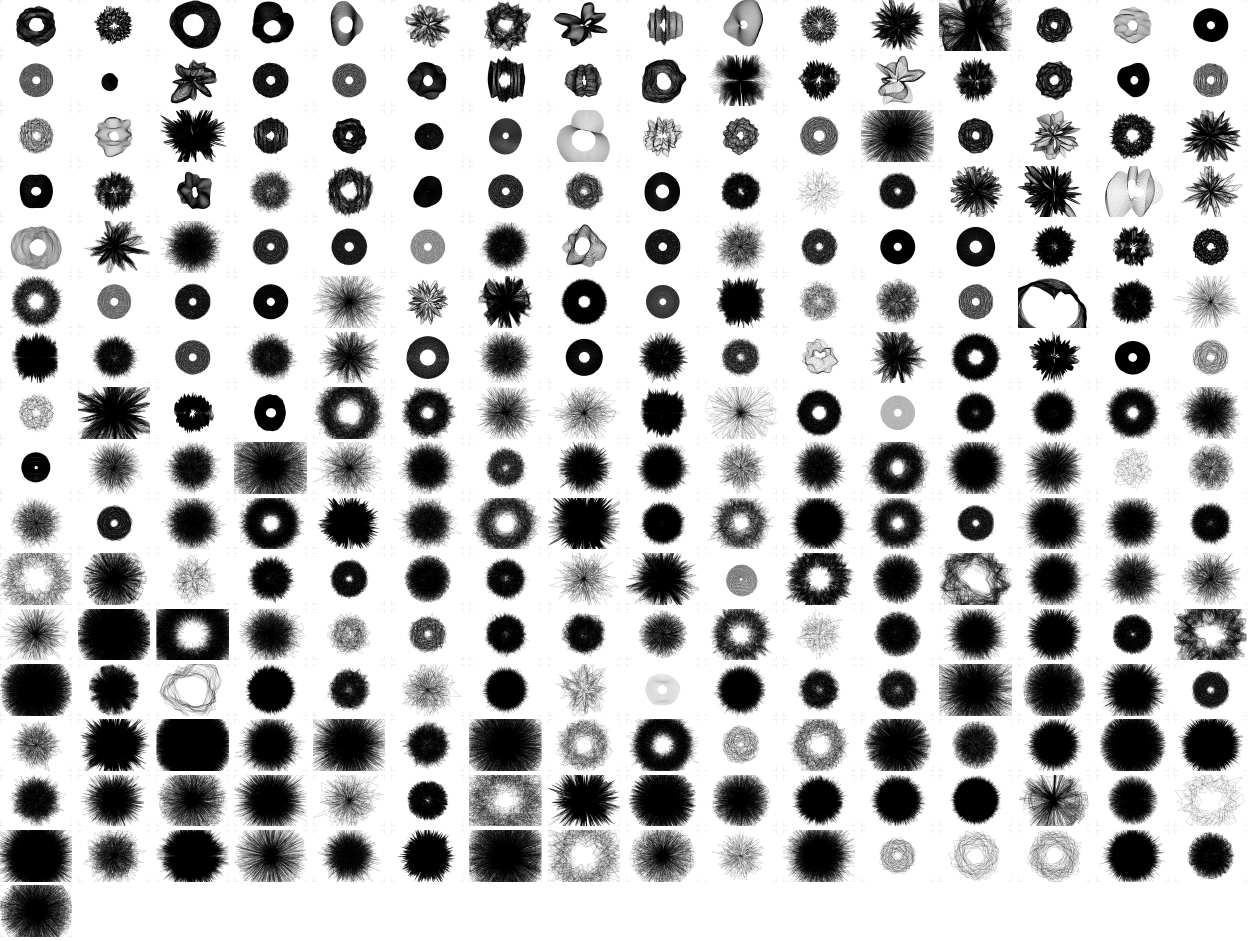} \\
    \includegraphics[width=0.45\textwidth]{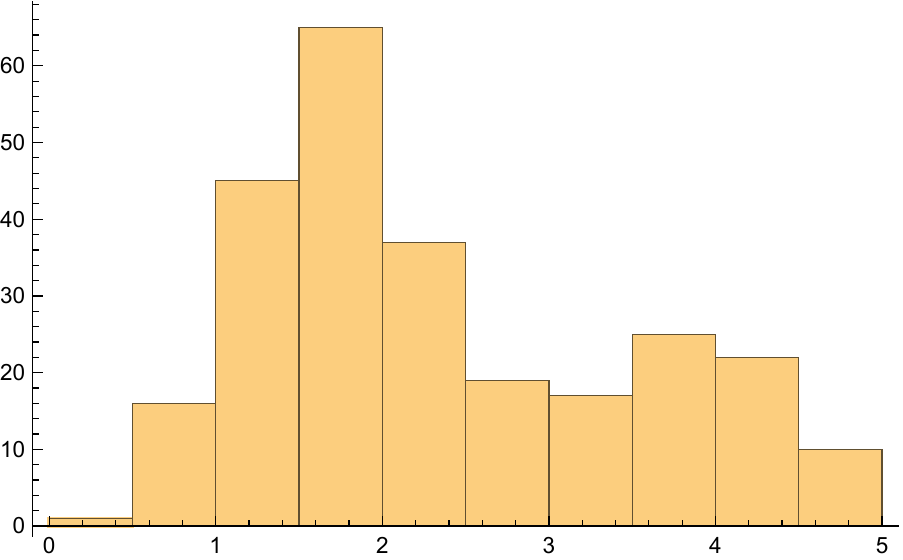} &
    \includegraphics[width=0.45\textwidth]{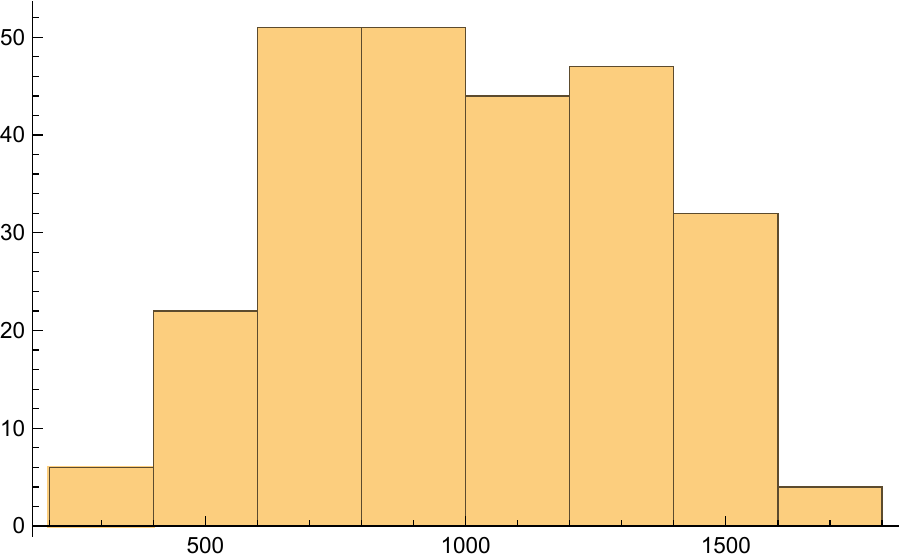} \\
    \end{tabular}
    \caption{Comparison of personal aesthetic evaluation of the same dataset using direct numeric score (left) and pairwise comparison (right). The images are shown in descending order from highest fitness at the top left. Histograms of the distribution are shown below the ordered images.}
    \label{fig:ratings}
\end{figure}

For the direct scoring method the artist looked at each drawing individually (presented in random order) and assigned each a real-valued numeric score from 0 (least appealing) to 5 (most appealing). For the pairwise comparison ranking process we used a browser-based application where pairs of images are displayed and compared by asking the user to answer the question ``which one of these images do you like the most?''. Comparing two images this way is the equivalent of a tournament (a battle for aesthetic superiority) where the possible outcome at each round (comparison) is one of `win', `loose' or `draw'. Images were ranked using an implementation of the Glicko ranking system \cite{glickman1998glicko, glickman1999parameter} based on their performance against other images. The ranking system gives each player (image) a score that becomes more accurate proportionate to the number of times it is part of a tournament outcome. The system also includes a rating deviation ($RD$) measure that represents the reliability of a player's rating. We performed sufficient comparisons to ensure that every image had an $RD < 250$, giving a high confidence in the ranking. The direct score ranking was performed after the pairwise comparison ranking. The results of these two methods of personal aesthetic ranking are shown in Fig. \ref{fig:ratings}.

As the figure shows, there is some difference in the ordering, even though the same person was performing both rankings on the same dataset. This is likely due to a number of factors, including the close visual similarity of many of the random phenotypes (making differentiation difficult), fatigue (over 1000 comparisons were evaluated), and the imprecise nature of aesthetic judgment. Nonetheless, the two rankings have a Spearman rank correlation coefficient of $r=0.74$ with a $p$-value $< 10^{-4}$, indicating a good consistency between the ranking methods.

\begin{figure}
    \centering
    \begin{tabular}{c|c}
     \includegraphics[width=0.48\textwidth]{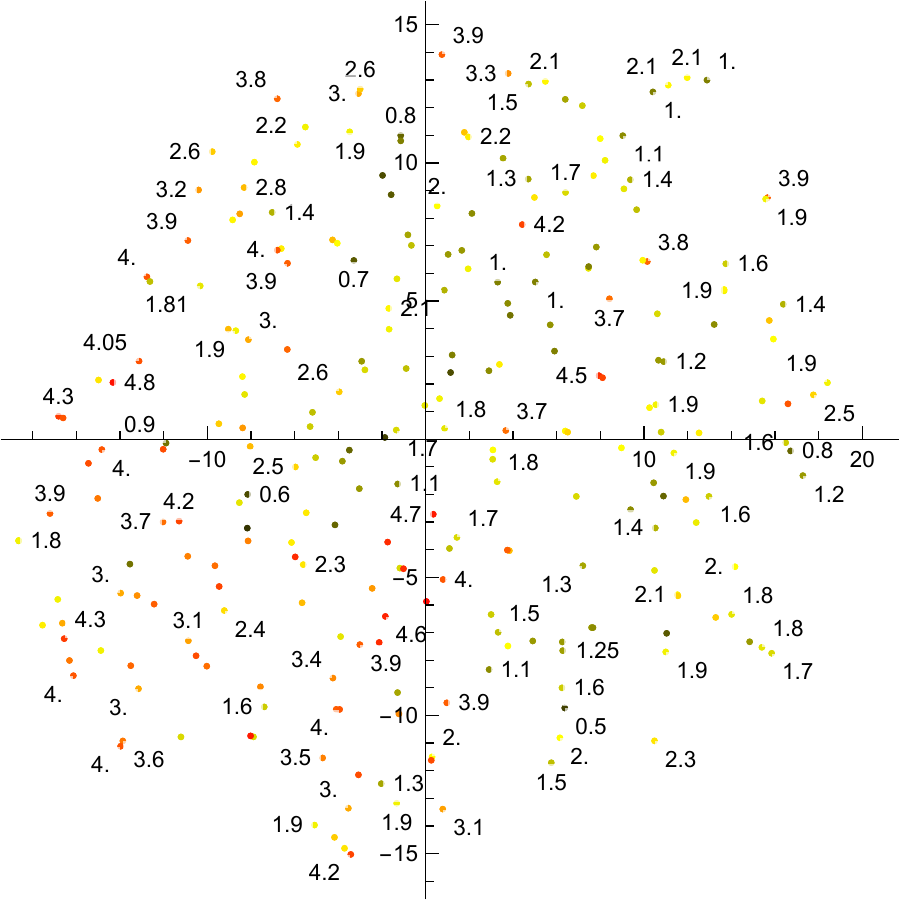} &
     \includegraphics[width=0.52\textwidth]{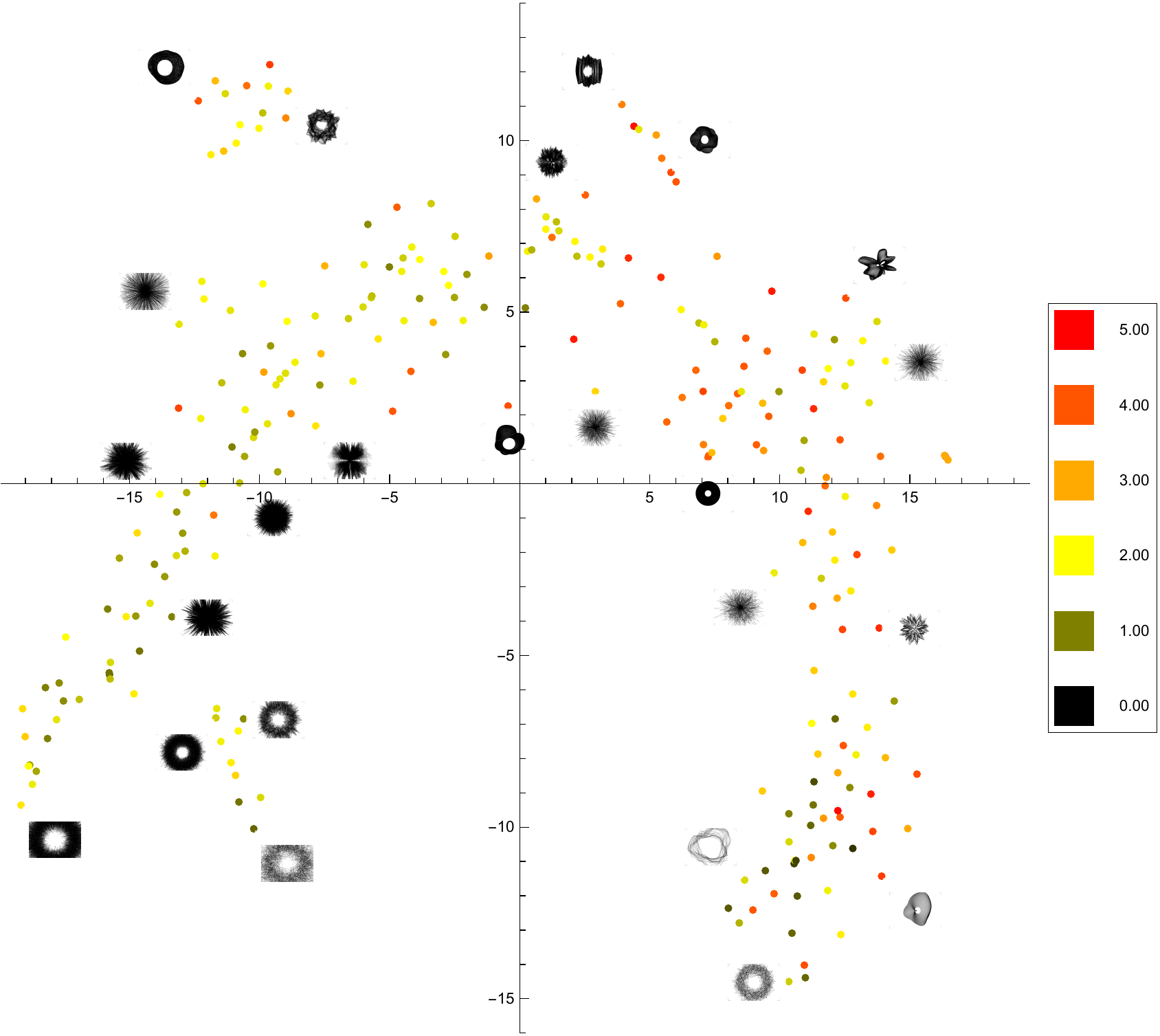} \\
    \end{tabular}
    \caption{Plots showing dimension reduced features (using the TSNE method) in geneotype space (left) and phenotype space (right). Each point is colour coded based on aesthetic rating as per the legend. Selected points show the image (phenotype) that corresponds to the point.}
    \label{fig:spaces}
\end{figure}

To further understand the relationship between genotype space, phenotype space and aesthetic measure we generated two visualisations (Fig. \ref{fig:spaces}). The visualisation on the left shows the genotype space dimensionally reduced from 17 dimensions to 2 dimensions using the t-SNE dimension reduction method \cite{maaten2008visualizing}, with a perplexity of 15 and $\epsilon = 10$. Each point on the visualisation represents an individual genotype, and the point is coloured according to its aesthetic score as shown in the legend on the far right of the figure (the numerical rating is also shown next to each point in the diagram). As can be seen, the random sampling produces a uniformly distributed scattering of points and while there is a slight increase in higher fitness individuals in the lower left quadrant, there are no obvious regions of high aesthetic fitness. The visualisation on the right shows the same dataset, but this time in phenotype space. To compute the 2D position of each phenotype we again dimensionally reduced each image (rasterised to a resolution of $1024 \times 768$ pixels) using the t-SNE algorithm (perplexity $= 15$ and $\epsilon = 10$) and colour coded each phenotype based on its aesthetic score. In this visualisation some structure can be observed, with many high-fitness individuals located along the right-hand `horseshoe' scattering of phenotypes, with a high concentration in the upper right quadrant. Notice also a smaller, isolated patch in the top upper-left quadrant. This visualisation shows that the phenotype space does have structure and that certain regions can be considered the `sweet spots' or `Klondike spaces', where high fitness individuals are more likely to be found. Additionally, the visualisation also shows that such regions are not unique, and there are multiple regions of high fitness and supporting the idea that there is no single best phenotype, just different regions of high quality, but visually distinct individuals.

\subsubsection{Complexity and Morphological Measures}

The same dataset of images were also evaluated using the image feature metrics described in \cite{McCormackEnigma2021}) and statistical measures of image intensity (mean, variance, centroid). The measures tested included image entropy and energy, morphological Euler number, algorithmic and structural complexity, fractal dimensional and fractal aesthetic measures (see \cite{McCormackEnigma2021} for details on each measure). These measures were calculated for each image and then compared to the artist-assigned rankings. The results of this comparison showed that the highest correlation was achieved by the skewness of an image's histogram (skew), and with the mean image intensity value (both with $r=0.54$, p-value $< 10^-3$). 

Based on these results, we selected mean pixel value ($\mu$) as an approximation for fitness, since this measure demonstrated the equal highest correlation and is also computed efficiently. However, the obvious flaw in this measure is that the highest fitness individual is an all-white image, so we compensate by defining a `hat' function with maximum fitness at the point $\mu = \gamma$:
\begin{equation}
     F(I) = \begin{cases}
        (\gamma - \mu_{min}) \mu(I) + \mu_{min} & \text{if } \mu_{min} \le \mu(I) < \gamma \\
        (\mu_{max} - \gamma) \mu(I) - \gamma & \text{if } \gamma \le \mu(I) \le \mu_{max} \\
        0 & \text{otherwise}
        \end{cases}
        \label{eqn:fitness}
\end{equation}
\noindent
where $F(I)$ is the fitness of an image, $I$; $\mu(I)$ is the normalised mean intensity of $I$, $\mu_{min}$ and $\mu_{max}$ are points for minimum and maximum intensity boundaries for a non-zero fitness value (eliminating cases that are effectively all black or all white). For our system we use the values $\mu_{min} = 0.05$, $\mu_{max} = 0.95$ and $\gamma = 0.75$.
Having devised a reasonable fitness measure, we then looked at how to measure diversity in a population of phenotypes using our system.

\subsection{Diversity}
\label{ss:diversity}

\begin{figure}
    \centering
     \includegraphics[width=\textwidth]{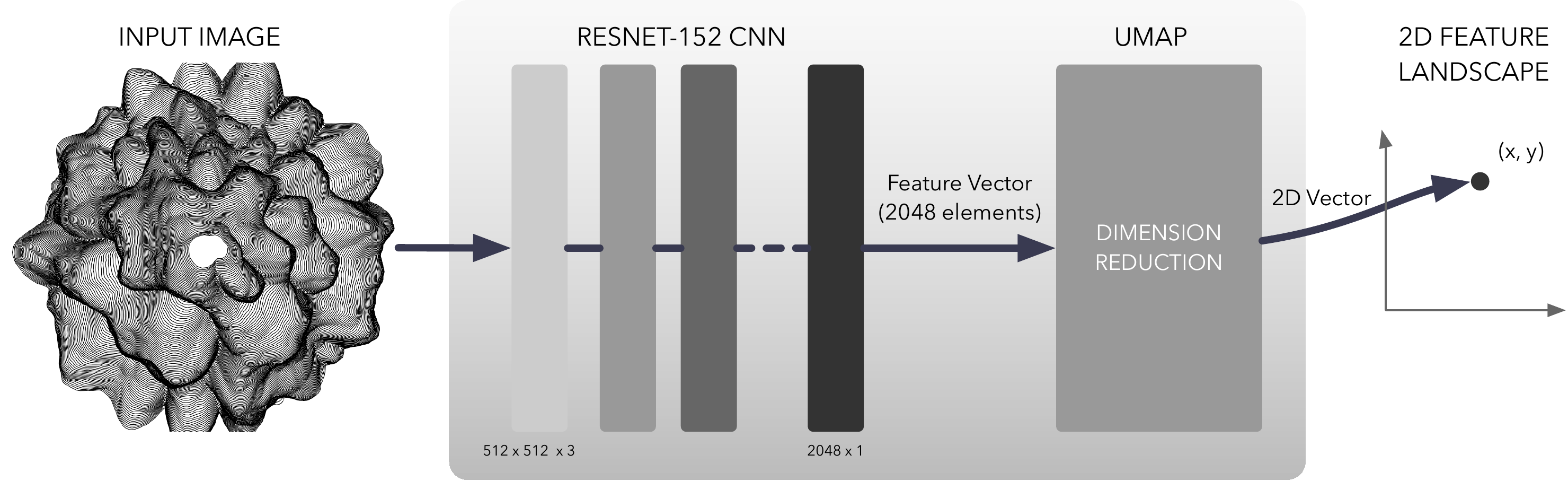}
    \caption{Evaluating diversity: each image is fed into a CNN classifier (ResNet-152), with the last 4 layers removed, leaving a 2048 element feature vector as the network's output. This is then dimensionally reduced using a UMAP algorithm, giving a 2-dimensional feature space vector.  }
    \label{fig:CNN-UMAP}
\end{figure}

\begin{figure}
    \centering
     \includegraphics[width=0.8\textwidth]{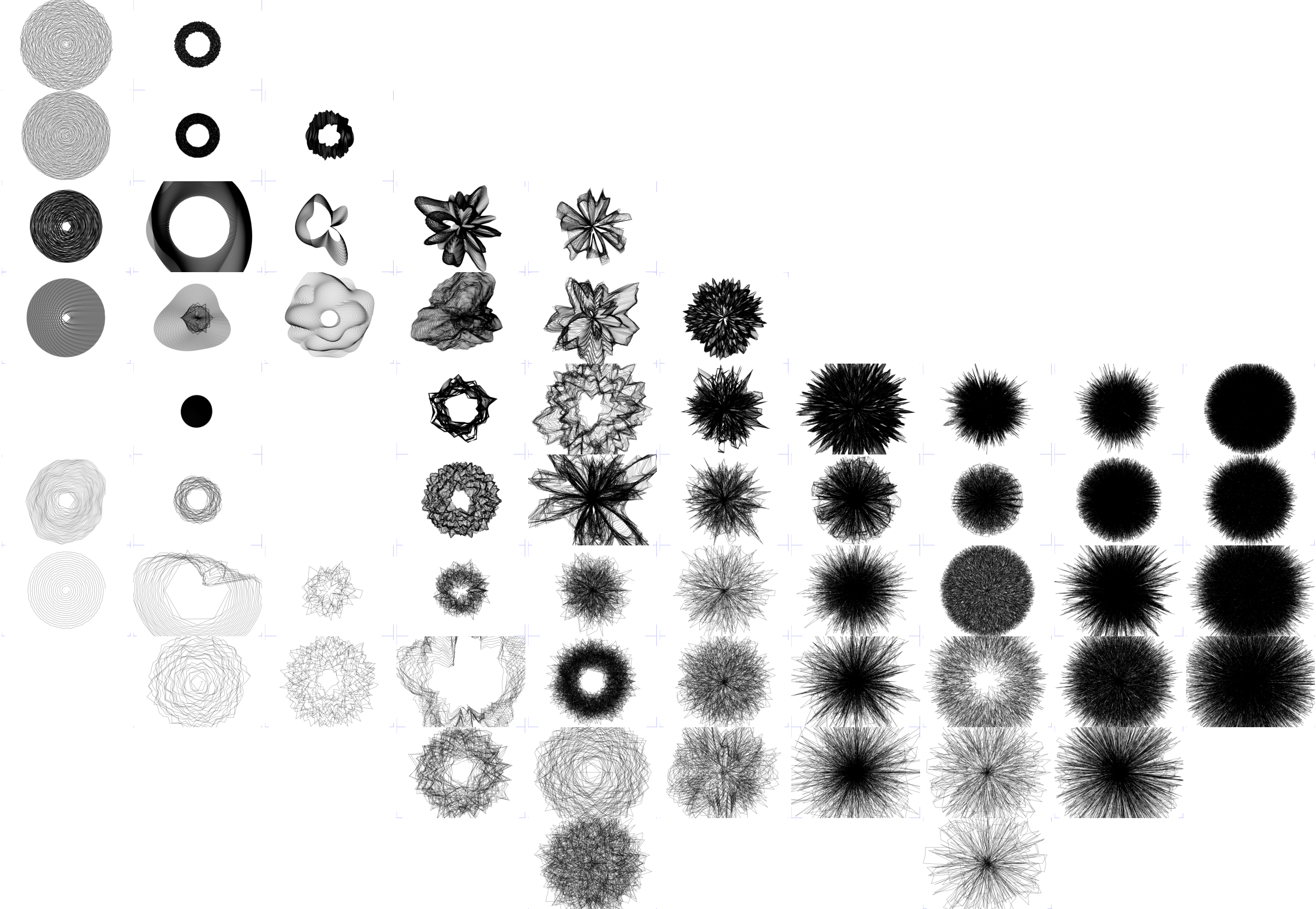}
    \caption{Sample of the dataset used to train the UMAP model classified into a $10 \times 10$ grid. }
    \label{fig:classification}
\end{figure}

To map the system's design space -- i.e. to visualise and understand the phenotypical differences between the drawings that it is capable of producing -- we used a combination of a convolutional neural network (CNN, a modified version of ResNet152 trained on the ImageNet database of real images \cite{deng2009imagenet}) for feature extraction, and Uniform Manifold Approximation and Projection (UMAP) \cite{mcinnes2018umap} for dimension reduction. Combining these methods we are able to classify generated images on a 2D plane, based on their visual characteristics (Fig. \ref{fig:CNN-UMAP}). This 2D space is then `quantised' into a grid, where each cell represents a different visual class of object (Figure \ref{fig:classification}). 

In order to define boundaries for the design landscape, we trained the CNN + UMAP model with a set of 738 randomly generated images, which includes the 257 images used for the manual aesthetic evaluation. Images are fed into the CNN as vectors of shape $(w, h, \mathbf{c})$, where $w$ and $h$ are the width and height of the image in pixels and $\mathbf{c}$ represents the $R$, $G$ and $B$ channels of each image. The CNN returns a 2048 element feature vector for each image. 
This vector is then reduced to 2-dimensions using UMAP, which gives us a reasonable estimation of the boundaries of the latent search space of our generative system, which we use to perform the dimensional reduction of newly generated images.

\subsection{MAP-Elites Implementation}
\label{ss:mapelites}

Our implementation of quality-diversity search is based on MAP-Elites \cite{mouret2015illuminating}. We use the measures of fitness and diversity introduced in sections \ref{ss:fitness} and \ref{ss:diversity} to evolve an elite population of drawings from our generative system.

The algorithm is initialised by creating an empty 2D \emph{feature space,} quantised into a grid $s = (n \times n)$ cells. The total number of generations ($e$) and population size at each generation ($\lambda$) is defined.

At each generation, a random cell from the feature space is sampled. If the cell is empty, a population of randomly generated drawings is created. Otherwise, the drawing in the cell is used as a parent for the new generation, which is created using a simple mutation procedure, where we define mutation rate $r$ -- the probability of an allele ($g_i$) to be mutated -- and mutation factor $f$ -- the maximum variation of a mutated allele.

Once a generation is created, we use \textit{.png} versions of the svg drawings to perform pixel-based evaluation, which gives us the fitness of each drawing (Eqn. \ref{eqn:fitness}), as well as their locations in the feature space. A drawing will be placed on the feature space grid if a) its corresponding cell is empty, or b) its fitness is higher than the one of the drawing occupying the cell. If a drawing is placed on the grid, it is also preserved in an archive.

The process repeats until the feature space grid has been updated for the desired number of generations.

\section{Experiments and Results}
\label{s:exp_results}

Our experiments were devised to test the effectiveness of the proposed approach as support for creative discovery. The goal was to evolve a \textit{landscape} of alternatives, where all the drawings produced by our generative system met a predefined quality standard based on the fitness function described in section \ref{ss:fitness}. The setup for these initial tests consisted of the following initial parameters:  $s = (8 \times 8)$, $e = 100$, $\lambda = 25$, $r = 0.25$ and $f = 0.15$.

\begin{figure}[t]
    \centering
     \includegraphics[width=0.7\textwidth]{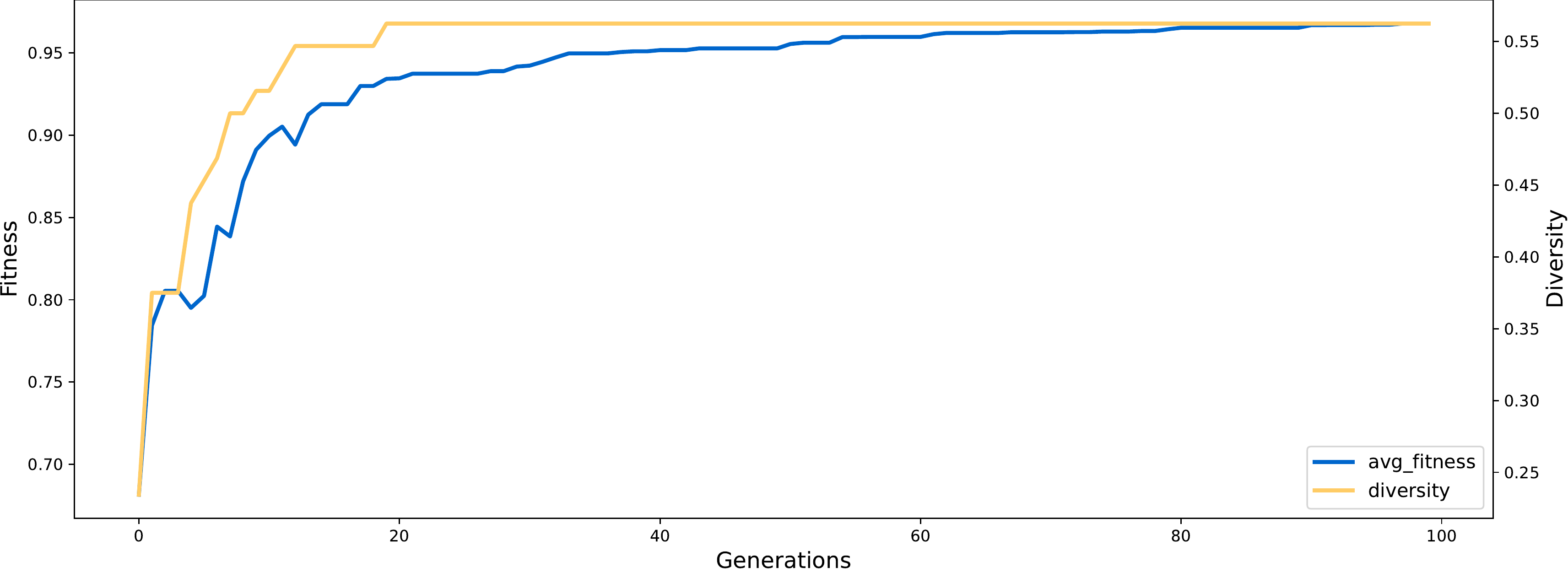}
    \caption{Time series of evolutionary run showing mean population fitness (blue) and population diversity (yellow)}
    \label{fig:fitness-diversity}
\end{figure}

The trajectory of the evolutionary process can be observed in Fig. \ref{fig:fitness-diversity}. Fitness is calculated as the mean fitness of all the drawings in the feature space. The ratio of cells in the grid that are populated is used as the population diversity measure. This trajectory shows how the feature space improves at different rates in both aspects. Diversity increases quickly, peaking at generation 60. Fitness, on the other hand, exhibits a slightly slower progression, where sudden increases in diversity produce slight setbacks that are overcome over a few generations.

\begin{figure}[!b]
    \centering
    \begin{tabular}{cc}
         \includegraphics[width=0.49\textwidth]{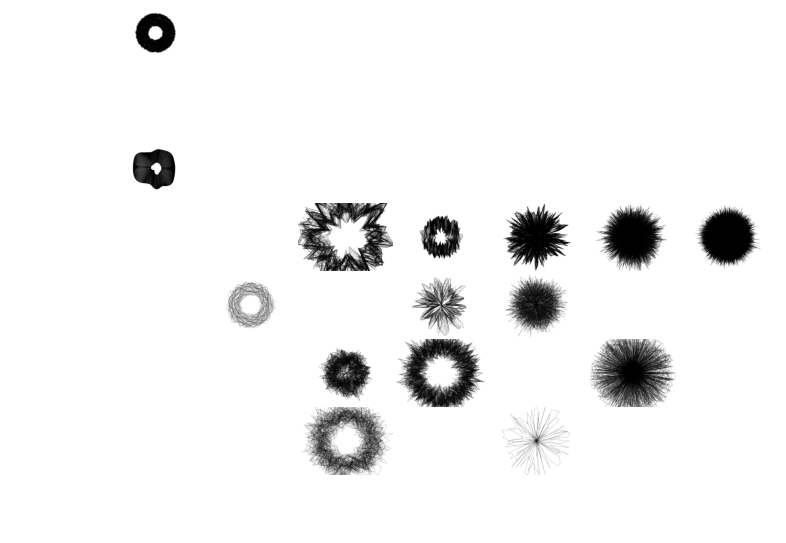} & 
         \includegraphics[width=0.49\textwidth]{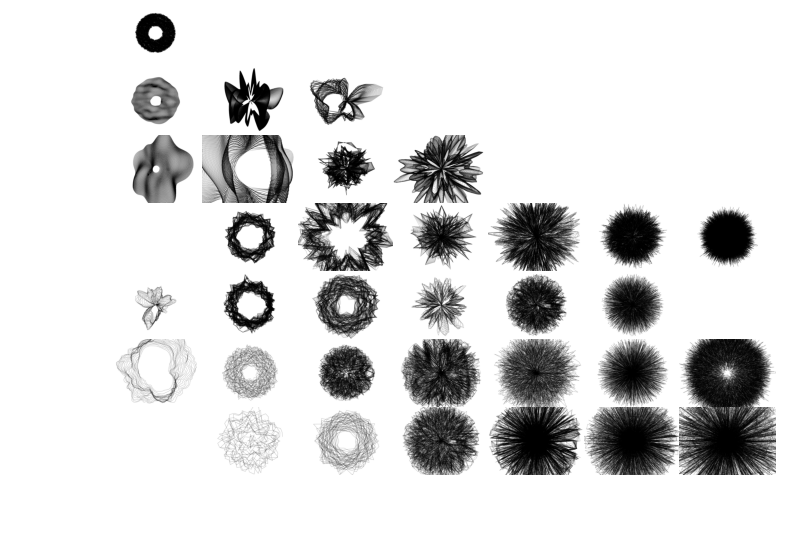} \\
         a) Grid at generation 0 & b) Grid at  generation 9 \\
         \includegraphics[width=0.49\textwidth]{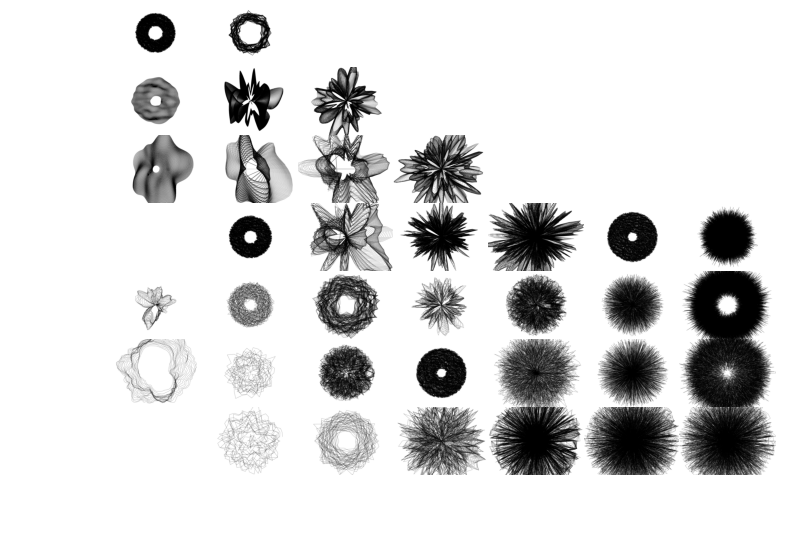} & 
         \includegraphics[width=0.49\textwidth]{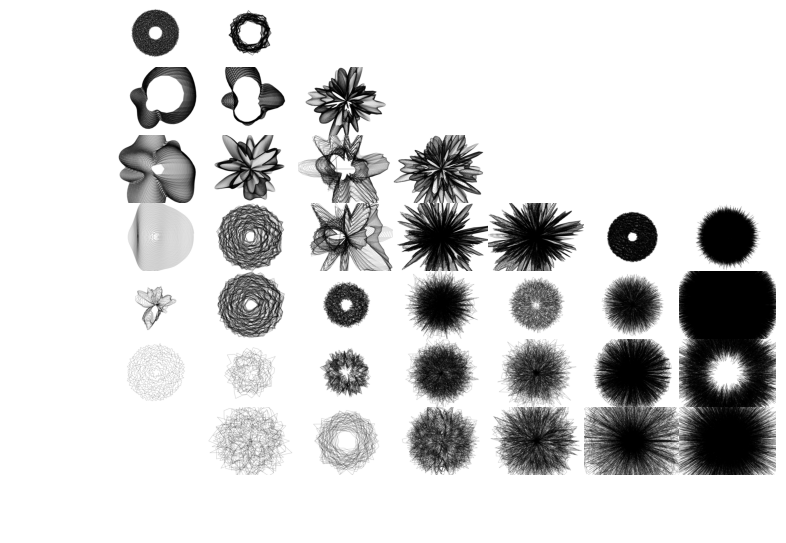} \\
         c) Grid at  generation 19 & d) Grid at generation 99
    \end{tabular}
     
    \caption{Examples of the elite population at four significant steps in the evolutionary process}
    \label{fig:grid-timeseries}
\end{figure}

Figure \ref{fig:grid-timeseries} shows the state of the feature space grid at significant stages of the evolutionary process. The image on the top left (a) shows the initial population placed on the feature space. Images b and c show the state of the feature space grid after increases in diversity, at generations 9 and 19. Finally, the image on the bottom right (d) shows the final elite population, at generation 99.

\begin{figure}[!b]
    \centering
    \includegraphics[width=0.7\textwidth]{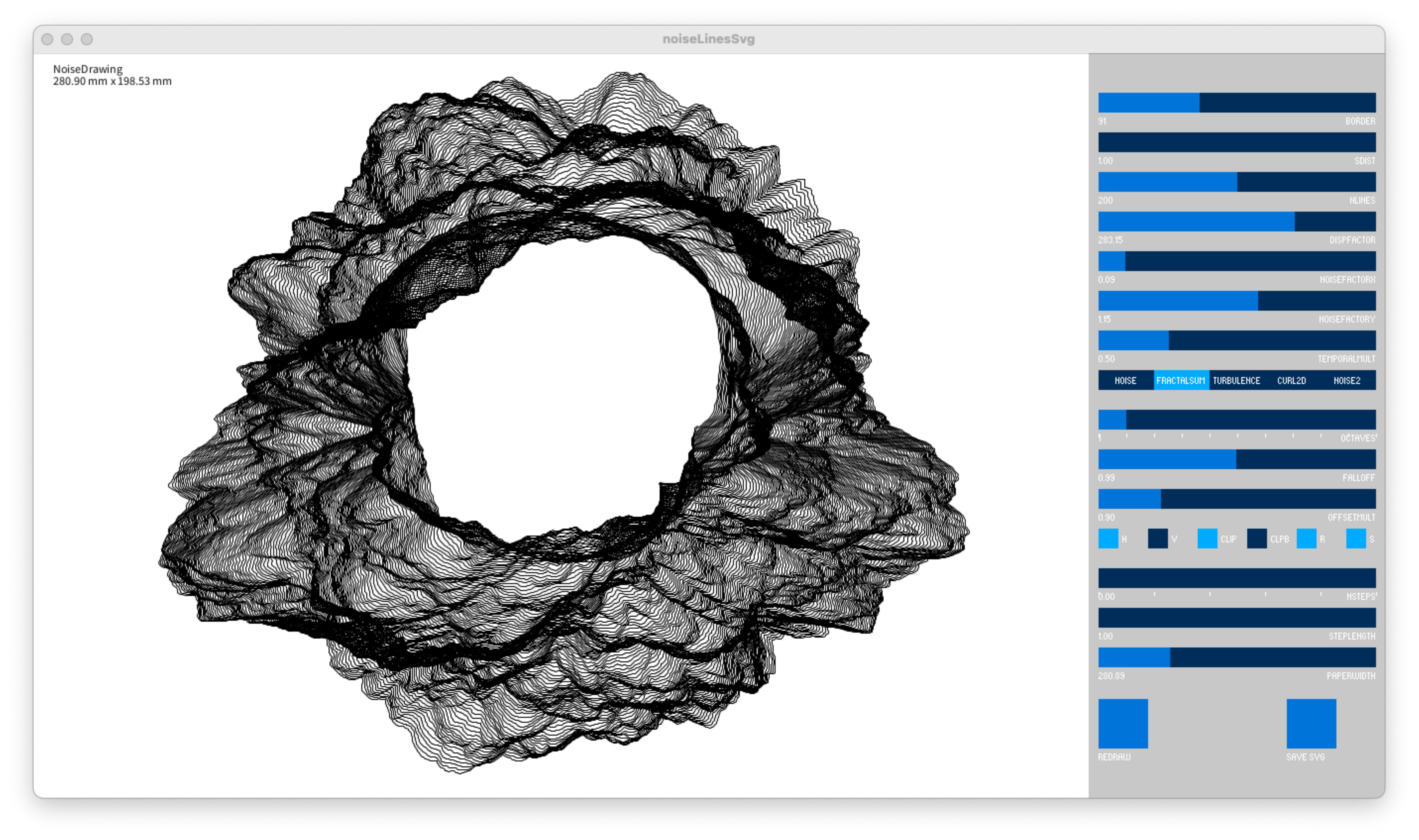}
    \caption{Screen shot of the interactive version of the generative system, with controls for each allele of the genome on the right.}
    \label{fig:interactive}
\end{figure}

\begin{figure}
    \centering
    \includegraphics[width=\textwidth]{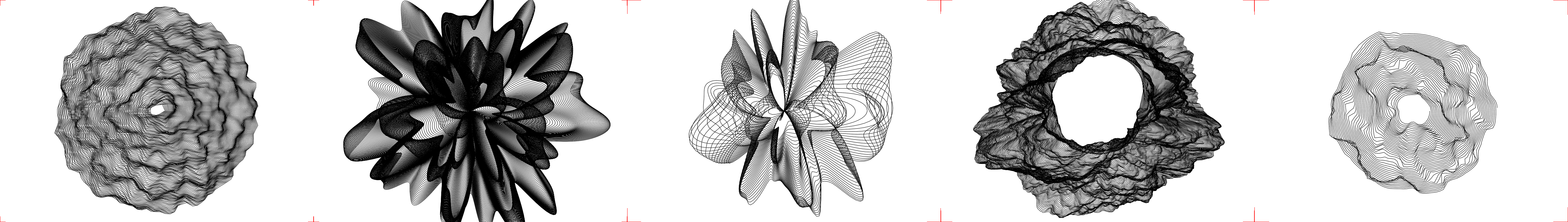}
    \caption{Example high fitness phenotypes found using the QD-search system.}
    \label{fig:bestOf}
\end{figure}

We ran the system several times and observed that the QD-search was consistently able to find a diverse range of high-fitness individuals. We compared these with forms the artist was able to find using an interactive version of the system that allows real-time manipulation of individual alleles in the genotype (Fig. \ref{fig:interactive}). Some of the forms the QD-search was able to find were highly unusual, certainly not readily apparent from many hours of exploring using the interactive system. Fig. \ref{fig:bestOf} shows some examples of forms found using the QD-search method described in this paper, demonstrating the system is able to find high fitness individuals that are diverse in appearance.

\section{Discussion}
\label{s:discussion}

As our results demonstrate, QD-search provides an interesting way for artists and designers to explore the creative possibilities of a generative system. As shown in Fig. \ref{fig:classification}, the visual display of diversity can assist human artists in understanding the visual possibilities that the system is capable of. Indeed, we had no idea the system could generate certain phenotypes that were found by QD-search, despite exploring the design space manually for several months.

While our results are promising, there are a number of limitations in the current study.
Firstly, for fitness we used a simple function based on the mean intensity of the image, as this had the equal highest correlation with the artist assigned perception of aesthetic quality for this specific system.
Clearly a more sophisticated fitness measure could give better nuances in high fitness individuals. Nonetheless, we were surprised at how well just using the function shown in Eqn. \ref{eqn:fitness} worked. However, such simple measures would be unlikely to generalise to other systems, so further work is needed to find high quality aesthetic fitness measures that work across different visual forms and styles.

Secondly, our results are limited to a single system and further research is needed to apply QD-search methods to a wider variety of creative systems. The fundamental challenge is in devising suitable computable measures of both quality and diversity that work for a specific generative system. However, the broad principles we have applied make this easy to test. Prior research has shown that while there exists reasonable computable proxies for specific aesthetic preferences \cite{McCormackEnigma2021}, the specific measure varies between system and individual, requiring the kind of correlation evaluation we performed in Section \ref{ss:fitness}. This requires significant human time, as the individual artist must manually rank or compare a sufficient number of random phenotypes to determine the best correlation with a variety of possible computable measures. Nonetheless, if the artist is willing to spend time to provide this information, the payoff can be significant. In our implementation QD-search is fully automated and doesn't require any further human evaluation once the initial fitness measure has been established. 

\section{Conclusion}
\label{s:conclusion}
We have demonstrated the use of QD-search methods in finding high fitness individuals in a generative art system. We began by generating a random sample of possible phenotypes and performing an artist-assigned measure of aesthetic quality on each. From this sample data we developed a fitness measure specific to the artist and the generative system, with good correlation between the computed measure and human-assigned scores. To compute a diversity measure we used a widely available CNN to provide a visual feature vector that was then dimensionally reduced from 2048 dimensions to two dimensions. Having both computable fitness and diversity measures allowed us to run a MAP-Elites QD-search and to find a series of diverse, high value individuals from the generative system.

While our results are currently limited to this single system, there are good reasons to believe that QD-search can be a valuable tool for artists and designers working with generative evolutionary systems. For future work we plan on testing the value of QD-search on more complex generative systems and on non-visual phenotypes such as sound synthesizers. The 2D layout of the QD-search also allows for interaction with the human designer, for example to clear certain cells and re-evolve the model for a specific diversity measure, or to click on an empty cell to direct the system to search for phenotypes with specific diversity measures.
%
%
%
\bibliographystyle{splncs04}
\bibliography{refs}
\end{document}